\documentclass[letterpaper, 10 pt, journal, twoside]{IEEEtran}
\IEEEoverridecommandlockouts

\usepackage[utf8]{inputenc} 
\usepackage[T1]{fontenc}    
\usepackage{hyperref}       
\usepackage{url}            
\usepackage{nicefrac}       
\usepackage{microtype}      
\usepackage[english]{babel}
\usepackage{graphicx}
\usepackage{amsmath}
\usepackage{amssymb}
\usepackage{booktabs}
\usepackage{mathtools}
\graphicspath{{figures/}} 
\usepackage{amsmath,amssymb,amsfonts}
\usepackage{algorithmic}
\usepackage{graphicx}
\usepackage{textcomp}
\usepackage{multirow}
\usepackage{adjustbox}
\usepackage{xcolor}
\usepackage{bm}
\usepackage{soul}
\usepackage{booktabs}
\usepackage{epstopdf}
\usepackage{cleveref}
\usepackage{makecell}
\usepackage{duckuments}
\usepackage{comment}
\usepackage[normalem]{ulem}
\usepackage{xcolor}
\usepackage[noadjust]{cite}
\usepackage{algorithm}
\usepackage{diagbox} 
\usepackage{multirow}
\usepackage{array}
\usepackage{tikz}
\usetikzlibrary{positioning,arrows.meta,shapes,fit}
\usetikzlibrary{shapes.geometric, arrows.meta, positioning, calc} 
\usetikzlibrary{backgrounds}

\IEEEpubidadjcol 
\usepackage[top=57pt, left=48pt, right=48pt, bottom=43pt]{geometry}
\def\IEEEtitletopspace{3pt}

\begin{document}
\title{\Large TCB-VIO: Tightly-Coupled Focal-Plane Binary-Enhanced Visual Inertial Odometry}
\author{
Matthew Lisondra$^{1\dagger}$, 
Junseo Kim$^{2\dagger}$, 
Glenn Takashi Shimoda$^{1}$,
Kourosh Zareinia$^{3}$, 
and Sajad Saeedi$^{4}$
\thanks{
© 2025 IEEE.  Personal use of this material is permitted.  Permission from IEEE must be obtained for all other uses, in any current or future media, including reprinting/republishing this material for advertising or promotional purposes, creating new collective works, for resale or redistribution to servers or lists, or reuse of any copyrighted component of this work in other works.
This work was supported by the Natural Sciences and Engineering Research Council of Canada (NSERC). We would like to thank Piotr Dudek, Stephen J. Carey, and Jianing Chen at the University of Manchester for kindly providing access to SCAMP-5.
}
\thanks{
$^{\dagger}$Authors contributed equally to the paper, 
$^{1}$University of Toronto, 
$^{2}$Delft University of Technology, 
$^{3}$Toronto Metropolitan University, 
$^{4}$University College London. 
%
Project website: \href{https://sites.google.com/view/tcb-vio}{https://sites.google.com/view/tcb-vio}}
}


\maketitle
\begin{abstract}
Vision algorithms can be executed directly on the image sensor when implemented on the next-generation sensors known as focal-plane sensor-processor arrays (FPSP)s, where every pixel has a processor. FPSPs greatly improve latency, reducing the problems associated with the bottleneck of data transfer from a vision sensor to a processor. FPSPs accelerate vision-based algorithms such as visual-inertial odometry (VIO). However, VIO frameworks suffer from spatial drift due to the vision-based pose estimation, whilst temporal drift arises from the inertial measurements. FPSPs circumvent the spatial drift by operating at a high frame rate to match the high-frequency output of the inertial measurements. In this paper, we present TCB-VIO, a tightly-coupled 6 degrees-of-freedom VIO by a Multi-State Constraint Kalman Filter (MSCKF), operating at a high frame-rate of 250 FPS and from IMU measurements obtained at 400 Hz. TCB-VIO outperforms state-of-the-art methods: ROVIO, VINS-Mono, and ORB-SLAM3.
\end{abstract}

\section{Introduction}
Agile and mobile robotic systems often operate under strict power and processing constraints. As a result, there is an increasing demand for low-latency and low-power camera technologies~\cite{falanga2020dynamic}, particularly for applications involving visual 3D pose estimation~\cite{ieeep_2018}. 
State-of-the-art algorithms utilizing conventional cameras typically achieve frame rates of 40–80 frames per second (FPS)~\cite{bloesch2017iterated, qin2018vins, campos2021orb}. These cameras rely on an architectural design where data is captured, digitized, and transferred to separate digital processing hardware for further computation. This traditional architecture, characterized by the modular separation of sensors and processors, can introduce significant latency and increase power consumption. On-sensor vision processing presents a novel paradigm that addresses these challenges by co-locating sensors and processors on the same chip. Focal-plane sensor processor arrays (FPSPs)~\cite{Zarandy} exemplify this approach, with each camera pixel equipped with a small processor capable of processing and sharing data with adjacent pixels (See Fig.~\ref{fig:fpsp}). This tight integration of sensing and processing reduces power consumption and minimizes delays caused by communication overhead, making FPSPs highly suitable for real-world robotic applications~\cite{Dudek_ScienceRob_2022},~\cite{lisondra2025embodied}.

Due to the limited space on FPSPs, per-pixel memory is highly constrained, and processing is typically performed in analog. 
Analog computation has inherent limitations, including reduced numerical precision, circuit inaccuracies, thermal effects, and noise leakage~\cite{Amant_ISCA}. These constraints challenge implementing computer vision algorithms and can lead to noisy results. Integrating additional sensing modalities~\cite{kim2024inverse}, such as inertial measurement units (IMU)s, using visual-inertial odometry (VIO) algorithms~\cite{scaramuzza2020aerial} can reduce the impact of the noise. However, adapting VIO algorithms to FPSPs requires a redesign of the processing pipeline to preserve the high-speed and low-power advantages of FPSPs.

\tikzset{
  block/.style={draw, thick, rectangle, minimum height=1cm, minimum width=2cm, align=center},
  smallblock/.style={draw, thick, rectangle, minimum height=1cm, minimum width=1cm, align=center},
  bus/.style={thick, line width=2mm, line cap=round}
}

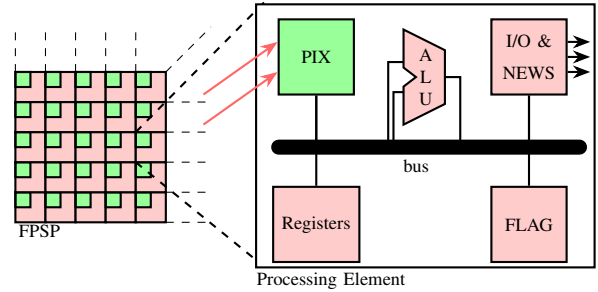
\begin{figure}
\centering
\begin{tikzpicture}[node distance=1.8cm]

\colorlet{pixcol}{green!40}
\colorlet{pecol}{red!20}

\begin{scope}[shift={(0,0)}, scale=0.4]
\def\N{5}\def\s{1}
\foreach \i in {0,...,\numexpr\N-1} {
  \foreach \j in {0,...,\numexpr\N-1} {
    \fill[fill=pecol] (\i*\s,\j*\s) rectangle ++(\s,\s);
    \fill[fill=pecol] ($(\i*\s+0.5*\s,\j*\s+0.5*\s)$) rectangle ++(0.5*\s,0.5*\s);
    \fill[fill=pixcol] ($(\i*\s,\j*\s+0.5*\s)$) rectangle ++(0.5*\s,0.5*\s);
    \draw[thick] ($(\i*\s,\j*\s+0.5*\s)$) rectangle ++(0.5*\s,0.5*\s);
    \draw[thick] (\i*\s,\j*\s) rectangle ++(\s,\s);
  }
}

\def\N{5}     
\def\s{1cm}    
\def\ext{1.5}

\foreach \j in {0,...,\numexpr\N-1} {
  \draw[dashed, black] (\N*\s, \j*\s) -- ++(\ext*\s, 0);
}

\foreach \i in {0,...,\numexpr\N-1} {
  \draw[dashed, black] (\i*\s, \N*\s) -- ++(0, \ext*\s);
}

\draw[dashed, black] (\N*\s,\N*\s) -- ++(\ext*\s,\ext*\s);

\draw[dashed, thick] (4,3) -- (8.3,7.35);
\draw[dashed, thick] (4,2) -- (8.3,-1.5);
\node at (0.8,-0.3) {\scriptsize{FPSP}};
\end{scope}

\begin{scope}[shift={(4, 2.2)}]

\fill[white!20] (-0.8,-2.8) rectangle (3.7,.7);
\draw[thick] (-0.8,-2.8) rectangle  (3.7,.7);

\node[smallblock, fill=pixcol] (pixel) {\scriptsize PIX};

\node[smallblock, fill=pecol, right=of pixel] (io) {\scriptsize I/O \&\\\scriptsize NEWS};

\node[smallblock, fill=pecol, below=1.2cm of io] (flag) {\scriptsize FLAG}; 

\node[smallblock, fill=pecol, below=1.2cm of pixel] (reg) {{\scriptsize Registers}};

\draw[bus] (-0.5, -1.2) -- (3.5, -1.2);

\node[rectangle,
      below right=-0.8cm and 1.2cm of pixel,
      rotate=-90, scale=0.55,
      minimum width=2cm, minimum height=1cm,
      draw=none, fill=none, inner sep=0pt] (alu) {};

\coordinate (aluSouthUp)   at ($(alu.south)+(0,0.2)$); 
\coordinate (aluSouthDown) at ($(alu.south)+(0,-0.2)$);

\begin{scope}[shift={(alu.center)}, rotate=-90, scale=0.55]

  \def\H{1.0}   
  \def\Wt{1.3}  
  \def\Wb{2.3}

  \def\npos{0.50}  
  \def\nfrac{0.20} 
  \def\ndepth{0.30} 

  \coordinate (A) at (-\Wt/2,\H/2);
  \coordinate (B) at ( \Wt/2,\H/2);
  \coordinate (C) at ( \Wb/2,-\H/2);
  \coordinate (D) at (-\Wb/2,-\H/2);

  \pgfmathsetmacro{\tOne}{\npos - 0.5*\nfrac}
  \pgfmathsetmacro{\tTwo}{\npos + 0.5*\nfrac}
  \coordinate (B1) at ($(D)!\tOne!(C)$);
  \coordinate (B2) at ($(D)!\tTwo!(C)$);
  \coordinate (M)  at ($(B1)!0.5!(B2)$);

  \coordinate (Tip) at ($(M) + (0,\ndepth)$);

  \path[fill=pecol, draw=black, thick]
    (A) -- (B) -- (C) -- (B2) -- (Tip) -- (B1) -- (D) -- cycle;
\end{scope}

\node [below right=-.75cm and .73cm of pixel, align=center] 
{\raisebox{0ex}{\scriptsize A}\\
\raisebox{0.8ex}{\scriptsize L}\\
\raisebox{1.6ex}{\scriptsize U}};

\node [below right=0cm and -1.5 of reg] {\scriptsize{Processing Element}};
\node [below right=-1.5cm and .45 of reg] {\scriptsize{bus}};

\draw[thick] (pixel.south) -- ++(0,-0.25) -- ++(0,-0.5);
\draw[thick] (alu.north) -- ++(0.2,0) -- ++(0.0,-1.0);
\draw[thick] (aluSouthUp) -- ++ (-0.2,0.0) -- ++(0.0,-1.1);
\draw[thick] (aluSouthDown) -- ++ (-0.13,0.0) -- ++(0.0,-0.7);
\draw[thick] (io.south) -- ++(0,-0.25) -- ++(0,-0.5);

\draw[thick] (reg.north) -- ++(0,0.5) -- ++(0,0.7);
\draw[thick] (flag.north) -- ++(0,0.5) -- ++(0,0.7);

\draw [red!60, thick, arrows = {-Stealth}] (-1.5,-.5) -- ($(pixel.west)+(0,.2)$);
\draw [red!60, thick, arrows = {-Stealth}] (-1.5,-.9) -- ($(pixel.west)+(0,-.2)$);

\draw [black, thick, arrows = {-Stealth}] ($(io.east)+(0,.2)$) -- ++ (0.3,0);
\draw [black, thick, arrows = {-Stealth}] ($(io.east)+(0,0)$) -- ++ (0.3,0);
\draw [black, thick, arrows = {-Stealth}] ($(io.east)+(0,-.2)$) -- ++ (0.3,0);

\end{scope}
\end{tikzpicture}
\vspace{-2mm}
\caption{
In a focal-plane sensor-processor array, each pixel, i.e., processing element, combines a photosensor circuit (PIX) with on-pixel compute resources, including an arithmetic-logic unit (ALU), input/output (I/O) circuits, local communication links (NEWS), local memory (Registers), and activity control (FLAG). 
This architecture enables image processing to be performed directly on the sensor~\cite{Dudek_ScienceRob_2022}.
}
\vspace{-6 mm}
\label{fig:fpsp}
\end{figure}

Early VIO algorithms were filtering-based, using the Extended Kalman Filter (EKF)~\cite{scaramuzza2020aerial}, \cite{bloesch2017iterated}, \cite{tanskanen2015semi}, Unscented Kalman Filter (UKF)~\cite{wan2000unscented}, 
and the popular MSCKF~\cite{mourikis2007multi}, where landmark positions are marginalized from the state vector to reduce the computational complexity. 
Later, smoothing approaches emerged, optimizing states over a window~\cite{leutenegger2013keyframe}, \cite{forster2016manifold} 
with tools like GTSAM~\cite{dellaert2012factor}, leading to variants such as VINS-Mono~\cite{qin2018vins}, SVO 2.0~\cite{forster2016svo}, Kimera VIO~\cite{rosinol2020kimera}, 
and ORB-SLAM3~\cite{campos2021orb}. 
Despite improved accuracy, they can suffer from inconsistencies and linearization errors~\cite{hesch2014camera}, \cite{dong2011motion}, \cite{huang2011observability}. 
VIO methods can also be categorized as loosely coupled or tightly coupled~\cite{scaramuzza2020aerial}. Loosely coupled methods, such as MSF~\cite{Lynen_IROS_2013_MSF}, process the two modalities independently and fuse them later. In contrast, tightly coupled algorithms, such as OKVIS~\cite{leutenegger2013keyframe}, jointly optimize visual and inertial data to achieve higher accuracy and robustness. For conventional cameras, however, the additional computation required by tightly coupled VIO often reduces update rates and increases the latency between image capture and state update~\cite{delmerico2018benchmark}. FPSPs can mitigate this trade-off: their in-pixel parallel processing enables fast on-sensor computations, while their ultra-high frame rate shortens the delay between image captures. 
For FPSPs, several visual odometry algorithms have been proposed~\cite{murai2023bit, boseVisualOdometryPixel2017}. To date, the only VIO algorithm is BIT-VIO~\cite{bitvio}, which is a loosely coupled method; no tightly coupled VIO approaches exist. This gap is addressed in this paper.

This paper presents the first tightly-coupled 6 degrees-of-freedom (DoF) VIO algorithm, designed for FPSPs. The algorithm coined, TCB-VIO, is a \textbf{T}ightly-\textbf{C}oupled VIO, utilizing \textbf{B}IT-enhanced features~\cite{murai2023bit}. 
TCB-VIO processes high-speed, on-sensor binary edge images and feature maps using a novel binary-enhanced Kanade-Lucas-Tomasi (KLT) tracker. It is an extended and modified adaptation of the tightly coupled framework, OpenVINS~\cite{Geneva2020ICRA}. The contributions of this paper are: 
1) First high frame-rate tightly-coupled VIO for FPSPs achieving a high frame rate of 250 FPS. 
2) 
The framework processes binary edge data and feature coordinates directly on-sensor, reducing computational cost. These outputs are then processed by the novel binary-enhanced KLT tracker at high frame rates.
3) Real-world evaluations comparing against ROVIO~\cite{bloesch2017iterated}, 
VINS-Mono~\cite{qin2018vins}, and 
ORB-SLAM3~\cite{campos2021orb}. 

In the rest of the paper, 
Sec.~\ref{sec:background} describes the background of FPSP. 
Sec.~\ref{sec:method} explains the TCB-VIO algorithm. 
Sec.~\ref{sec:results} details our experimental results. 
Sec.~\ref{sec:conc} concludes the work.

\section{Background: SCAMP-5 FPSP}
\label{sec:background}
A traditional camera consists of a 2D array of light-sensitive pixels. In contrast, FPSPs integrate a processor within each pixel on the same chip~\cite{Zarandy} (See Fig.~\ref{fig:fpsp}). FPSPs are also known as processor-per-pixel arrays (PPA) or cellular-processor arrays (CPA). One example of FPSP is the SCAMP-5 system~\cite{carey2013100}, which features a 256$\times$256-pixel FPSP. Each pixel, or Processing Element (PE), includes a photosensor circuit (PIX) and compute resources. These PEs operate in parallel, executing identical instructions on local data in a Single Instruction, Multiple Data (SIMD) architecture. Each PE can also communicate with its north, east, west, and south neighbors, shown as NEWS in Fig~\ref{fig:fpsp}. 
Each PE contains $7$ analog registers, $13$ 1-bit registers, and an ALU composed of 176 transistors, which means they can perform logical and arithmetic operations on the sensor. 
Arithmetic operations are performed directly in the analog domain using analog registers. Avoiding digitization improves the computational speed but introduces computational noise~\cite{carey2013100}. 
The computational results can be outputted in various formats, e.g. events, binary frames, or analog frames. 
The SCAMP-5 FPSP has been used in some robotics applications. 
Examples are 
target tracking~\cite{greatwood2017tracking}, 
agile drones~\cite{mcconville2020visual}, ~\cite{greatwood2018perspective},\cite{greatwood2019towards}, 
fast inference~\cite{liu2021direct, so2024pixelrnn}, \cite{auke}, \cite{stow2020cain}, \cite{Wong2020AnalogNetCN},
obstacle avoidance~\cite{chen2020proximity}, and 
localization~\cite{castillo2021weighted}, visual odometry~\cite{murai2023bit}, \cite{bitvio}, \cite{boseVisualOdometryPixel2017}, \cite{fourdofcamtracking},
and 
navigation~\cite{stow2022compiling}. 
It is worth noting that the emergence of event cameras has led to a new generation of VIO algorithms operating at high effective frame rates~\cite{vidal2018ultimate}, \cite{zihao2017event, rebecq2017real, vidal2018ultimate, mueggler2018continuous}
. Event cameras are promising for VIO in high-speed and low-light scenarios. However, unlike FPSPs, they lack the flexibility of user-programmable processing directly on the focal plane, which enables customized computation.

\section{Proposed Method}\label{sec:method}

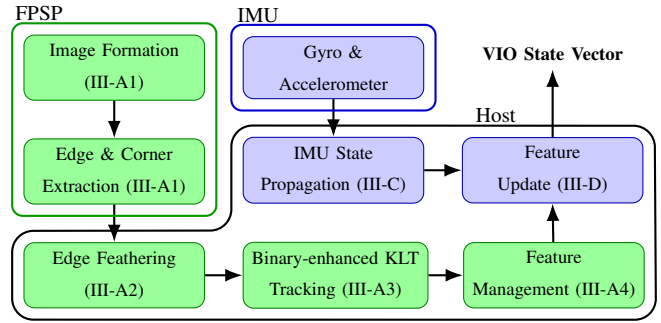
\begin{figure}[t!]
\centering
\begin{tikzpicture}[
    >=Latex,
    node distance=1.5cm,
    blockgreen/.style={
        rectangle, rounded corners, minimum width=2.4cm, minimum height=.6cm,
        draw=green!50!black, top color=viscol, bottom color=viscol, text centered
    },
    blockorange/.style={
        rectangle, rounded corners, minimum width=2.4cm, minimum height=.6cm,
        draw=blue!70!black, top color=imucol, bottom color=imucol, text centered
    },
    container/.style={
        rounded corners, draw=black, thick, inner sep=0.15cm
    }
]
\colorlet{viscol}{green!40}
\colorlet{imucol}{blue!20}
\node[blockgreen, align=center] (imgform) {\scriptsize Image Formation\\\scriptsize (\ref{sec:features})};
\node[blockgreen, align=center, below=0.5cm of imgform] (edgeext) {\scriptsize Edge \& Corner\\\scriptsize Extraction (\ref{sec:features})};

\node[blockorange, align=center, right=0.5cm of imgform] (imu) {\scriptsize Gyro \& \\\scriptsize Accelerometer};

\node[blockgreen, , align=center, below=0.5cm of edgeext] (adj) {\scriptsize Edge Feathering\\\scriptsize (\ref{sec:feather})};

\node[blockgreen, align=center, right=0.5 cm of adj] (feather) {\scriptsize Binary-enhanced KLT\\\scriptsize Tracking (\ref{sec:klt})};

\node[blockgreen, align=center, right=0.46cm of feather] (klt) {\scriptsize Feature\\\scriptsize Management (\ref{sec:bittype})};

\node[blockorange, align=center, below=0.5cm of imu] (stateprop) {\scriptsize IMU State\\\scriptsize Propagation (\ref{sec:prop_imu})};
\node[blockorange, align=center, right=0.5cm of stateprop] (bitupdate) {\scriptsize Feature\\\scriptsize Update (\ref{sec:on_manifold})};

\node[above=.9cm of bitupdate] (viostate) {\scriptsize \bf VIO State Vector};

\draw[->, thick] (imgform) -- (edgeext);
\draw[->, thick] (edgeext) -- (adj);
\draw[->, thick] (adj) -- (feather);
\draw[->, thick] (feather) -- (klt);

\draw[->, thick] (imu) -- (stateprop);
\draw[->, thick] (stateprop) -- (bitupdate);
\draw[->, thick] (bitupdate) -- (viostate);

\draw[->, thick] (klt) -- (bitupdate.south);
 
\node[container, draw=green!60!black, fit=(imgform)(edgeext), label={[xshift=-1.cm,yshift=-0.1cm]\footnotesize FPSP}] {};
\node[container, draw=blue!80!black, fit=(imu), label={[xshift=-1.cm,yshift=-0.1cm]\footnotesize IMU}] {};
\draw[container, draw=black, rounded corners=8pt]
  ($(adj.north west)+(-0.15,0.15)$) --
  ($(feather.north west)+(-0.15,0.15)$) --
  ($(stateprop.north west)+(-0.15,0.15)$) --
  ($(bitupdate.north east)+(0.15,0.15)$) --
  ($(klt.south east)+(0.15,-0.15)$) --
  ($(adj.south west)+(-0.15,-0.15)$) --
  cycle
  node[anchor=north west, xshift=6cm, yshift=1.8cm] {\footnotesize Host};
\end{tikzpicture}
  \caption{ 
  TCB-VIO processing pipeline. Numbers inside each block correspond to the paper sections describing them. Green blocks denote visual processing: the FPSP performs on-sensor corner and edge extraction once an image is formed, and the extracted features are transferred to the host where a novel binary-enhanced KLT tracker ensures fast and robust tracking. Blue blocks denote inertial propagation and updates via MSCKF.
  } 
  \vspace{-5 mm}
\label{fig:overview}
\end{figure}

Fig.~\ref{fig:overview} shows an overview of the TCB-VIO pipeline. 
Edges and corner features are computed on the FPSP, then transferred to a host for further processing with a novel binary-enhanced Kanade-Lucas Tomasi (KLT) tracker. The results are fused with IMU measurements. 
The division of the workload between the FPSP and the host is dictated by the computational characteristics of each stage. Feature extraction is highly parallelizable across pixels and 
is well-suited to the SIMD architecture of the FPSP. 
In contrast, the data fusion and the optimization required for the tracker involve iterative and global computations that are not well-suited to the current FPSP architecture, and are therefore performed on the host.

\subsection{Visual Processing}\label{sec:visualproc}
This section details visual processing steps, shown as green-filled blocks in Fig.~\ref{fig:overview}.
\subsubsection{Image Formation and Feature Extraction}\label{sec:features}
The front-end visual processing of corner and binary features occurs directly on the SCAMP-5 FPSP. 
We use a modified version of FAST-16 implemented on the FPSP \cite{chen2017feature} for the front-end corner feature-part, where the processing of all pixels is done in parallel on FPSP~\cite{chen2018scamp5d}, allowing for high frame-rate outputting. The same is done for the binary edge feature detection on the FPSP~\cite{murai2023bit}, where the FPSP analog registers were used, computed in parallel, like with the corner features. 
The rest of the processing steps are described as follows.

The set of $N_p$ detected corners (see Table~\ref{tab:key_parameters} for the size) and edge features acquired on the FPSP is stored in two binary feature maps: the corner map \(\mathbf{M}_c\) and the edge map \(\mathbf{M}_e\). These maps are transmitted to the host device at a high frame rate of 250 FPS and are formally defined as 
$\mathbf{M}_c, \mathbf{M}_e \in \{0, 1\}^{256 \times 256},$ 
where \(\mathbf{M}_c(i, j) = 1\) indicates that a corner feature is present at pixel location \((i, j)\), and \(\mathbf{M}_e(i, j) = 1\) denotes the presence of an edge feature at that pixel.

\subsubsection{Edge Feathering}\label{sec:feather}
Once a binary edge image is received on the host, it is converted to an 8-bit grayscale image, where all edges take the value of 128. To enhance edge continuity and introduce smooth intensity transitions for improved feature tracking, a Gaussian-based feathering operation is applied to the grayscale image, generating \(\mathbf{M}_\text{feathered}\). The goal is to produce softly graded edge intensities that improve robustness in gradient-based feature processing.
We empirically determine a suitable value of variance of the Gaussian kernel \(\sigma_e\) (see Table~\ref{tab:key_parameters} for the value) that produces desirable feathering, preserving high intensity at the center of edges while introducing sufficient gradient smoothness for tracking stability.

\subsubsection{Binary-enhanced KLT Tracking}\label{sec:klt}

To enable efficient, high frame-rate visual tracking on binary data from the FPSP, we adapt the traditional KLT tracking algorithm~\cite{lucas1981iterative}, \cite{tomasi1991detection}, \cite{shi1994good} to operate on the feathered grayscale image. 
Unlike conventional KLT pipelines, which rely on dense intensity derivatives from natural images, our approach utilizes minimal yet structured gradients derived from binary edge maps to maintain visual tracking robustness at significantly reduced computational cost.

The binary-enhanced KLT utilizes both corners and feathered edges. Each corner pixel \((x_i, y_i)\) for which \(\mathbf{M}_c(x_i, y_i) = 1\) is selected as a tracking point, forming the set \(\mathbf{C} = \{(x_i, y_i)\}\). 

While the corner locations come from the binary corner map \(\mathbf{M}_c\), the actual intensity information used for tracking is drawn from \(\mathbf{M}_\text{feathered}\), which encodes smooth gradients generated from the edge map \(\mathbf{M}_e\). For each corner, a spatial window \(\mathbf{W}_i\) (see Table~\ref{tab:key_parameters} for the size) is centered on \((x_i, y_i)\), and optical flow is computed over these windows. The feathering process introduces local intensity uniqueness between adjacent edges, ensuring that similar binary edge structures become distinguishable for gradient-based motion estimation.

For each feature, the displacement \(\Delta \mathbf{u}\) is computed by minimizing the photometric error between consecutive frames:
\begin{align}
\sum_{(x, y) \in \mathbf{W}_i} \left( I_t(x, y) - I_{t+1}(x + \Delta u_x, y + \Delta u_y) \right)^2,
\label{eq:intensity_error_parallel}
\end{align}
where \(I_t\) and \(I_{t+1}\) are the feathered edge images at times \(t\) and \(t+1\), respectively. The optimization is solved iteratively, leading to  
$\Delta \mathbf{u}^k = \mathbf{H}^{-1} \mathbf{g}$, 
where~\cite{lucas1981iterative}:
\begin{align}
\mathbf{H} &= \sum_{(x, y) \in \mathbf{W}_i} \nabla I(x, y) \nabla I(x, y)^\top, \label{eq:hessian_parallel} \\
\mathbf{g} &= \sum_{(x, y) \in \mathbf{W}_i} \nabla I(x, y) \left(I_t(x, y) - I_{t+1}(x, y)\right). \label{eq:gradient_parallel}
\end{align}

The updated feature locations are given by 
$\mathbf{C}_{t+1} = \mathbf{C}_t + \Delta \mathbf{u}^k.$ 
Tracking continues until convergence or the maximum iteration count:
$\|\Delta \mathbf{u}^k\| < \epsilon,  \forall (x, y) \in \mathbf{C}.$ 
The binary-enhanced KLT tracker outputs a sequence of high-frequency 2D features with observations \(\mathbf{z}_i^k \in \mathbb{R}^2\), where feature \(i\) is tracked across a sliding window of camera frames indexed by \(k \in \{t_{i}, \dots, t_{i-N_c}\}\). 
These features serve as visual measurements, $\mathbf{z}_{\text{meas}}(\cdot)$, in the MSCKF backend to update the state (Sec.~\ref{sec:on_manifold}). Beforehand, a decision is made whether to include them in the state vector, as discussed next.

\subsubsection{Feature Management}\label{sec:bittype}
The MSCKF uses features to constrain camera poses during state estimation~\cite{mourikis2007multi}. These features are not explicitly maintained in the state vector and are therefore referred to as out-of-state (or MSCKF) features. Following Li and Mourikis~\cite{li2013optimization}, to improve robustness, a subset of features can be explicitly included in the filter state. These are called in-state (or SLAM) features, as they provide long-horizon geometric constraints. In TCB-VIO, the in-state features are stored in $\mathbf{x}_{BIT}$, described in Sec.~\ref{sec:vio_state}, and updated as described in Sec.~\ref{sec:on_manifold}.

Despite the high frame rate of the FPSP, careful management of SLAM (in-state) features is required, since considering all detected features as in-state would significantly increase both the state dimension and the computational cost of updates. To balance accuracy and efficiency, only a small, reliable subset of features is promoted to the state vector. This classification is based on spatiotemporal observability and track length: features with sufficiently long and stable tracks are designated as in-state, whereas shorter-lived tracks remain out-of-state~\cite{Geneva2020ICRA}.  
Features tracked for fewer than $N_c$ frames, defined in Sec.~\ref{sec:vio_state}, remain out-of-state; those tracked for at least $N_c$ are promoted to in-state, with $N_c=15$ in our experiment (Table~\ref{tab:key_parameters}).

\begin{figure}[t]
\centering
\begin{tikzpicture}[
    >=Latex,
    node distance=1.5cm,
    blockgreen/.style={
        rectangle, rounded corners, minimum width=3.0cm, minimum height=1cm,
        draw=green!50!black, top color=green!50, bottom color=green!80, text centered
    },
    blockorange/.style={
        rectangle, rounded corners, minimum width=3.0cm, minimum height=1cm,
        draw=orange!70!black, top color=orange!50, bottom color=orange!80, text centered
    },
    container/.style={
        rounded corners, draw=black, thick, inner sep=0.15cm
    }
]
    \centering
\node[anchor=south west, inner sep=0] (raw2) at (0,0)
{\includegraphics[width=0.2\columnwidth]{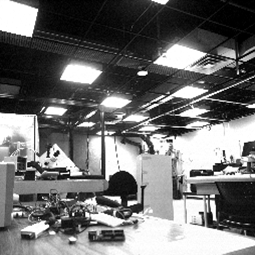}};

\node[below left= -1.5 cm and -2.5 cm of raw2 , inner sep=0] (raw1){\includegraphics[width=0.2\linewidth]{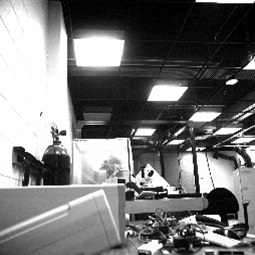}};
\node[below= .55cm of raw2 , inner sep=0] (edge2){\includegraphics[width=0.2\linewidth]{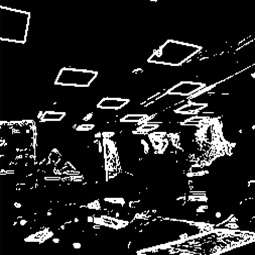}};

\node[below left= -1.5 cm and -2.5 cm of edge2, inner sep=0](edge1){\includegraphics[width=0.2\linewidth]{figures/edges2.png}};
\node[right= 1 cm of raw2 , inner sep=0] (corner2){\includegraphics[width=0.2\linewidth]{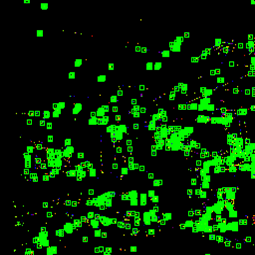}};
    
\node[below left= -1.5 cm and -2.5 cm of corner2, inner sep=0] (corner1){\includegraphics[width=0.2\linewidth]{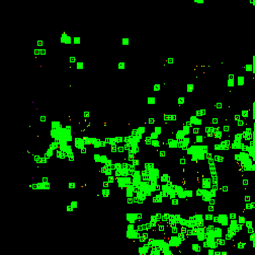}};
\node[right= 1 cm of edge2 , inner sep=0] (blur2){\includegraphics[width=0.2\linewidth]{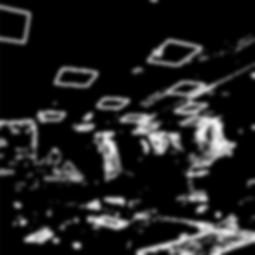}};

\node[below left= -1.5 cm and -2.5 cm of blur2, inner sep=0] (blur1){\includegraphics[width=0.2\linewidth]{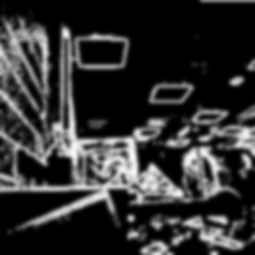}};
\node[below left= -1 cm and -5.9 cm of corner2, inner sep=0] (klt){\includegraphics[width=0.32\linewidth]{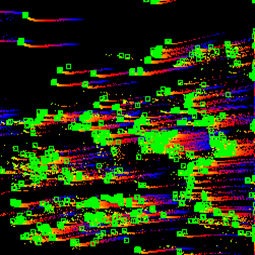}};
\begin{pgfonlayer}{background}
     \draw[->, thick, red, -stealth] (raw2) -- (edge2);
     \draw[->, thick, red, -stealth] (raw2) -- (corner2);
     \draw[->, thick, red, -stealth] (edge2) -- (blur2);
     \draw[->, thick, red, -stealth] (corner2) -- (klt);     
     \draw[->, thick, red, -stealth] (blur2) -- (klt);     
\end{pgfonlayer}   

\draw[container, draw=black, rounded corners=3pt]
  ($(klt.north west)+(-0.05,0.05)$) --
  ($(klt.north east)+(+0.05,0.05)$) --
  ($(klt.south east)+(+0.05,-0.82)$) --
  ($(blur2.south west)+(-0.05,-0.35)$) --  
  ($(blur2.north west)+(-0.05,0.05)$) --    
  ($(blur2.north east)+(1.22,0.05)$) --    
cycle
  node[anchor=north west, xshift=2.1cm, yshift=.4cm] {\scriptsize Host};
\node [below left=-.1 cm and -1.55 cm of klt] {\scriptsize{KLT Tracking}};
\node [below right=-.03 cm and .59 cm of blur2] {\scriptsize{Feathered Edges}};

\draw[container, draw=black, rounded corners=3pt]
  ($(raw2.north west)+(-0.05,0.05)$) --
  ($(corner2.north east)+(+0.75,0.05)$) --
  ($(corner2.south east)+(+0.75,-0.35)$) --
  ($(raw2.south east)+(+0.75,-0.35)$) --  
  ($(edge2.south east)+(+0.75,-0.35)$) --    
  ($(edge2.south west)+(-.05,-0.35)$) --
cycle
  node[anchor=north west, xshift=5.3cm, yshift=-.1cm] {\scriptsize FPSP};
\node [below left=-.1 cm and -.82 cm of raw2] {\scriptsize{Images}};
\node [below left=-.1 cm and -.79 cm of edge2] {\scriptsize{Edges}};
\node [below left=-.1 cm and -.84 cm of corner2] {\scriptsize{Corners}};

\end{tikzpicture}
\vspace{-5 mm}
\caption{Overview of the binary-enhanced KLT Tracking. The FPSP generates binary corners and edges. 
On the host, binary edges are feathered, and the KLT tracker operates in spatial windows centered on each corner feature.}

  \label{fig:bklt_overview}
\end{figure}

\subsection{Visual-Inertial State Vector}\label{sec:vio_state}
Sec.~\ref{sec:visualproc} described how features are extracted and processed.
This section describes the inertial/visual state vectors, followed by inertial/visual fusion in the remaining sections.
The notation used here is adopted from OpenVINS~\cite{Geneva2020ICRA}. 
The state vector $\mathbf{x}_i$ 
consists of all current estimates of the filter up to time $t_i$. It has four parts: 
the current navigation state $\mathbf{x}_{NAV}$, 
a set of $N_c \in \mathbb{N}$ IMU clones of a sliding window $\mathbf{x}_{IMU}$, 
a set of $N_r \in \mathbb{N}$ reprojected from 2D-to-3D mapped BIT-features $\mathbf{x}_{BIT}$ ({explained in more detail in Sec.~\ref{sec:bittype}}) and, a set of camera calibration intrinsics and extrinsics parameters $\mathbf{x}_{C}$:
\begin{align}
    \mathbf{x}_i &= \begin{bmatrix} \mathbf{x}_{NAV}^{\top} & \mathbf{x}_{IMU}^{\top} & \mathbf{x}_{BIT}^{\top} & \mathbf{x}_{C}^{\top} \end{bmatrix}^{\top}.
    \label{eq:state_vector}
\end{align}
The current IMU navigation state $x_{NAV}$ is defined as
\begin{align}
    \mathbf{x}_{NAV} &= \begin{bmatrix} {{}^{I_i}_G \mathbf{q}^{\top}} & {}^G \mathbf{p}_{I_i}^{\top} & {}^G \mathbf{v}_{I_i}^{\top} & \mathbf{b}_{\omega,i}^{\top} & \mathbf{b}_{a,i}^{\top} \end{bmatrix}^{\top},
\end{align}
where ${{}^{I_i}_G \mathbf{q}^{\top}}, {}^G \mathbf{p}_{I_i}^{\top}, \mathbf{v}_{I_i}^{\top}$ represents the quaternion orientation, position, velocity from the IMU-frame ($I_i$) at time $t_i$ to the global world-frame $G$, respectively. The $\mathbf{b}_{\omega,i}^{\top}, \mathbf{b}_{a,i}^{\top}$ are the gyroscopic and accelerometer biases.
The IMU clones is 
\begin{align}
    \mathbf{x}_{IMU} &= \begin{bmatrix} \mathbf{x}_{IMU_{t_i}}^{\top} & \cdots & \mathbf{x}_{IMU_{t_i-N_c}}^{\top} \end{bmatrix}^{\top},
    \label{eq:ximu}
\end{align}
\noindent where $\mathbf{x}_{IMU_{t_i}}^{\top}=\begin{bmatrix} {{}^{I_i}_G \mathbf{q}^{\top}} & {}^G \mathbf{p}_{I_i}^{\top} \end{bmatrix}$. Note that the indexing is from $t_i$ to $t_i-N_c$, forming a sliding window of states, from current to a past window of $N_c$ prior.
The extracted BIT features are reprojected in the global frame as 3D key features,
\begin{align}
    \mathbf{x}_{BIT} &= \begin{bmatrix} {}^G \mathbf{p}_{BIT,1}^{\top} & \cdots & {}^G \mathbf{p}_{BIT,N_r}^{\top} \end{bmatrix}^{\top},
    \label{eq:x_bit}
\end{align}
\noindent where ${}^G \mathbf{p}_{BIT,i}^{\top}=(x_i,y_i,z_i)$ is in global frame $G$. The frame-to-frame tracking is processed at 250 FPS using a novel binary-enhanced Kanade-Lucas-Tomasi (KLT) tracker, in which the binary edge images and feature maps are computed 
on-sensor.

Finally, between the camera-frame $C$ with respect to the IMU-frame $I_i$ at time $t_i$, the camera intrinsics and extrinsics,
\begin{align}
    \mathbf{x}_{C} &= \begin{bmatrix} {{}^{I_i}_C \mathbf{q_i}^{\top}} & {}^C \mathbf{p_i}_{I_i}^{\top} & {}^G \mathbf{\Phi_i}_{0}^{\top} \end{bmatrix}^{\top}
\end{align}
are continually optimized, where ${}^{I_i}_C \mathbf{q_i}^{\top}, {}^C \mathbf{p_i}_{I_i}^{\top}$ represents the quaternion orientation and position from the IMU-to-camera frame. ${}^G \mathbf{\Phi_i}_{0}^{\top}$ is the camera intrinsic vector that updates in time, in the global frame $G$.

\subsection{Propagation of State by IMU}\label{sec:prop_imu}

The IMU is taken as a 6-axis input, where measurements in the local IMU frame are
\begin{align}
    {}^I \bm{\omega}_{local}(t) &= {}^I \bm{\omega}_T(t) + \mathbf{b}_{G}(t) + \bm{n}_G(t), \label{eq:w_local}\\
    {}^I \bm{a}_{local}(t) &= {}^I \bm{a}_T(t) + {}^I_G \mathbf{R}(t) {}^G \bm{g} + \mathbf{b}_{A}(t) + \bm{n}_A(t), \label{eq:a_local}
\end{align}
where ${}^I \bm{\omega}_T(t), {}^I \bm{a}_T(t)$ are the true and un-noised gyroscope and accelerometer measurements. $\bm{b}_{G}(t), \bm{b}_{A}(t)$ represent the gyroscopic and accelerometer biases, respectively. ${}^I_G \mathbf{R}(t)$ is the rotation from global frame $G$ to IMU-frame, and is used in transforming the approximate estimated gravity vector $\bm{g}$ to the local IMU frame. $\bm{n}_G(t), \bm{n}_A(t)$ are  white Gaussian noises.

The IMU state evolves over time with notations adopted from \cite{trawny2005indirect} and kinematic propagation model from \cite{Geneva2020ICRA}, \cite{mourikis2007multi}:
\begin{align}
    {{}^{I}_G \mathbf{q}}(t) &= \frac{1}{2} \begin{bmatrix} 
    -\left[ {}^I \bm{\omega}_T(t) \times \right] & {}^I \bm{\omega}_T(t) \\ 
    -{}^I \bm{\omega}_T(t)^{\top} & 0 
    \end{bmatrix} \begin{bmatrix}
    {}^I \bm{\omega}_T(t) \\ 
    \bm{I}_t^G \bm{q}
    \end{bmatrix}, \\
    {}^G \dot{\mathbf{p}_{I}}(t) &= {}^G \mathbf{v}_I(t),~~~~\,
    {}^G \dot{\mathbf{v}}_I(t) = \bm{I}_t^G \mathbf{R}^{\top} {}^I \bm{a}_T(t), \\
    \dot{\mathbf{b}}_G(t) &= \bm{n}_{G}(t),~~~~~~~\dot{\mathbf{b}}_A(t) = \bm{n}_{A}(t).
\end{align}
The true un-noised gyroscope and accelerometer IMU measurements $\omega_T(t),\bm{a}_T(t)$ can be obtained by rearranging 
Eq.~\eqref{eq:w_local}-\eqref{eq:a_local}, respectively with raw input IMU measurements $\bm{\omega}_{local}(t), \bm{a}_{local}(t)$ in the local-frame.

The solutions of the kinematic equations start with~\cite{Geneva2020ICRA}
\begin{align}
    {}^{I_{i+1}}_G \mathbf{R} = \exp \left( -\int_{t_i}^{t_{i+1}} {}^I \bm{\omega}_T(Z) \, dZ \right) {}^{I_i}_G \mathbf{R}
\end{align}
Here, the rotation from global frame $G$ to IMU-frame from the last estimate ${}^{I_i}_G \mathbf{R}$ is updated incrementally. 
Then the position is acquired from the last position, velocity ${}^G \mathbf{p}_{I_i},{}^G \mathbf{v}$, respectively. involving a double-integral on the last acceleration $\mathbf{a}_T(Z)$:
\begin{align}
{}^G \mathbf{p}_{I_{i+1}} = & {}^G \mathbf{p}_{I_i} + {}^G \mathbf{v}_{I_i} (t_{i+1}-t_i) - \frac{1}{2} {}^G \mathbf{g} (t_{i+1}-t_i)^2 \notag \\
& + {}^{I_i}_G \mathbf{R}^{\top} \int_{t_i}^{t_{i+1}} \int_{t_i}^{s} {}^{I_i}_{I_Z} \mathbf{R}^{\top} {}^I \mathbf{a}_T(Z) \, dZ \, ds .
\end{align}
The velocity is then propagated incrementally at $t_{i+1}$ integrating the last acceleration $\mathbf{a}_T(Z)$:
\begin{equation}
{}^G \mathbf{v}_{I_{i+1}} =  {}^G \mathbf{v}_{I_i} + {}^{I_i}_G \mathbf{R}^{\top} \hspace{-2mm}\int_{t_i}^{t_{i+1}} \hspace{-5mm}{}^{I_i}_{I_Z} \mathbf{R}^{\top} {}^I \mathbf{a}_T(Z)dZ- {}^G \mathbf{g} (t_{i+1}-t_i). 
\end{equation}

The biases are also propagated by integrating the random walk noises $\bm{n}_{G}, \bm{n}_{A}$. They are zero-mean Gaussians:
\begin{equation}
\mathbf{b}_{G_{i+1}} = \mathbf{b}_{G_i} + \int_{t_i}^{t_{i+1}} \hspace{-5mm}\bm{n}_{G}(Z)dZ,~
\mathbf{b}_{A_{i+1}} = \mathbf{b}_{A_i} + \int_{t_i}^{t_{i+1}} \hspace{-5mm}\bm{n}_{A}(Z)dZ.
\end{equation}
The MSCKF backend of TCB-VIO linearizes the nonlinear IMU kinematic model and incrementally succeeds the state along with propagating the covariance, as explained in \cite{mourikis2007multi}.

\subsection{Update of 3D Point Features}\label{sec:on_manifold}
At each measurement time-step $t_m$, a tracked corner feature, described in Sec.~\ref{sec:klt}, is reprojected as a 3D bearing landmark, based on the following measurement model $\mathbf{h}(\cdot)$: 
\begin{align}
    \mathbf{z}(t_m) &= \mathbf{h}(\mathbf{x}(t_m)) + \mathbf{n}(t_m) \notag \\
    &= \mathbf{T}_d(\mathbf{T}_p(\mathbf{T}_t(\mathbf{T}_r(\boldsymbol{\lambda}, \boldsymbol{K}), \mathbf{x}_{CG})), \zeta) + \mathbf{n}(t_m),
    \label{eq:meas_model}
\end{align}
where $n(t_m)$ is zero-mean {Gaussian} noise, and $\mathbf{h}{(\cdot)}$ is a chain of functions following the standard projection pipeline~\cite{Geneva2020ICRA}:
\begin{itemize}
\itemindent=-10pt
    \item $\mathbf{T}_r(\cdot)$ converts the landmark parameterization $\lambda$ (inverse-depth or 3D position, with any context such as an anchor frame/extrinsics denoted by $\mathbf{K}$) into global coordinates. 
    \item $\mathbf{T}_t(\cdot, \mathbf{x}_{CG})$ transforms the global point to the current camera frame using the camera pose in the global frame ${G}$, $\mathbf{x}_{CG}$. 
    \item $\mathbf{T}_p(\cdot)$ orthogonally projects onto the normalized image plane. 
    \item $\mathbf{T}_d(\cdot,\zeta)$ maps normalized coordinates to distorted pixel coordinates using intrinsic distortion parameters $\zeta$. 
\end{itemize}
The predicted feature measurement is $\mathbf{z}_{\text{pred}}(t_m)=\mathbf{h}(\mathbf{x}(t_m))$. For the update, the residual is $r(t_m) =\mathbf{z}_{\text{meas}}(t_m) - \mathbf{h}(\mathbf{x}(t_m))$. 
For MSCKF features, residuals accumulated over the clone window are projected into the left null-space of the landmark Jacobian. This operation removes the landmarks analytically and allows for updating only the cloned poses. 

\section{Experimental Results}\label{sec:results}
This section details the experiments along with discussions. 

\subsection{Experimental Setup}
TCB-VIO operates with the SCAMP-5 FPSP, processing at 250. 
IMU measurements, at 400 Hz, and grayscale images for benchmarking are sourced from an Intel D435i RealSense. These two sensors are rigidly attached to a fixture. (See~\cite{web:tcbvio}) Benchmarks methods are ROVIO~\cite{bloesch2017iterated}, VINS-Mono~\cite{qin2018vins}, and ORB-SLAM3~\cite{campos2021orb} using grayscale and IMU data. TCB-VIO cannot be evaluated on standard benchmark datasets due to on-sensor computation. 
For indoor experiments, the ground truth is provided by a Vicon motion capture system running at 300 Hz. 
To synchronize and calibrate the FPSP and IMU, Kalibr~\cite{furgale2013unified} was used. 
For metrics, the {absolute trajectory error (ATE) and relative trajectory error (RTE) are stated as a median and root mean squared error (RMSE)~\cite{sturm2012benchmark}.}
13 trajectories were used for benchmarking (10 indoors and 3 outdoors). 
The trajectories are created by rapidly shaking the fixture in all directions to simulate violent motions.  
Trajectories \#1-3 are mostly rotational.
Trajectories \#4-7 are mostly translational. 
Trajectories \#8-10 are also mostly rotational, but more challenging than trajectories \#1-3, based on their average angular velocities (see Table~\ref{tb:ate_rte}, top row). 
Trajectories \#11-13 (see Table~\ref{tb:start_to_end}) are similar to trajectories \#1-3, but occurring outdoors. The length and duration of each trajectory are also provided in the tables.

\subsection{Parameters of TCB-VIO} \label{sec:optim}
Table~\ref{tab:key_parameters} lists the key parameters used in the experiments. An exhaustive search over a wide range tuned these. 
Graphs of the parameter search are on the project webpage~\cite{web:tcbvio}. While $N_r$, the size of $\mathbf{x}_{BIT}$ in Eq.~\eqref{eq:x_bit}, may grow over time, the number of landmarks used for updating SLAM and MSCKF is limited, denoted with $N_\text{SLAM, update}$ and $N_\text{MSCKF, update}$.
\begin{table}[t!]
\centering
\caption{Key parameters for TCB-VIO configuration.}
\begin{tabular}{l l}
\hline
\textbf{Description (Symbol)} & \textbf{Value} \\ \hline
Number of points extracted and tracked ($N_p$) & 800 \\ \hline
Max IMU clones in the sliding window, ($N_c$ at Eq.~\ref{eq:ximu}) & 15 \\ \hline
Max SLAM-represented BIT feat in update ($N_\text{SLAM, update}$) & 30 \\ \hline
Max MSCKF-represented BIT feat in update ($N_\text{MSCKF, update}$) & 60 \\ \hline
Gaussian feathering variance for edge smoothing ($\sigma_e$) & 2.5 \\ \hline
Spatial window size in pixels (\(\mathbf{W}_i\) at Eq.~\eqref{eq:intensity_error_parallel}) & $21 \times 21$ \\ \hline
\end{tabular}
\label{tab:key_parameters}
\end{table}
\subsection{Comparisons with ROVIO and VINS-Mono}
As shown in Table~\ref{tb:ate_rte}, TCB-VIO demonstrates superior robustness against fast and hostile motions, achieving the lowest ATE and RTE across all experiments, compared with ROVIO and VINS-Mono. 
VINS-Mono suffers from partial failures. ROVIO exhibits reasonable robustness, avoiding large errors throughout all experiments. However, its ATE and RTE values are still up to 4 to 10 times higher than those of TCB-VIO. Error values for partial failures are not reported, as they are at least $10\times$ larger than the lowest error.

\begin{figure}[t!]
\centering
  \includegraphics[width=1\columnwidth]{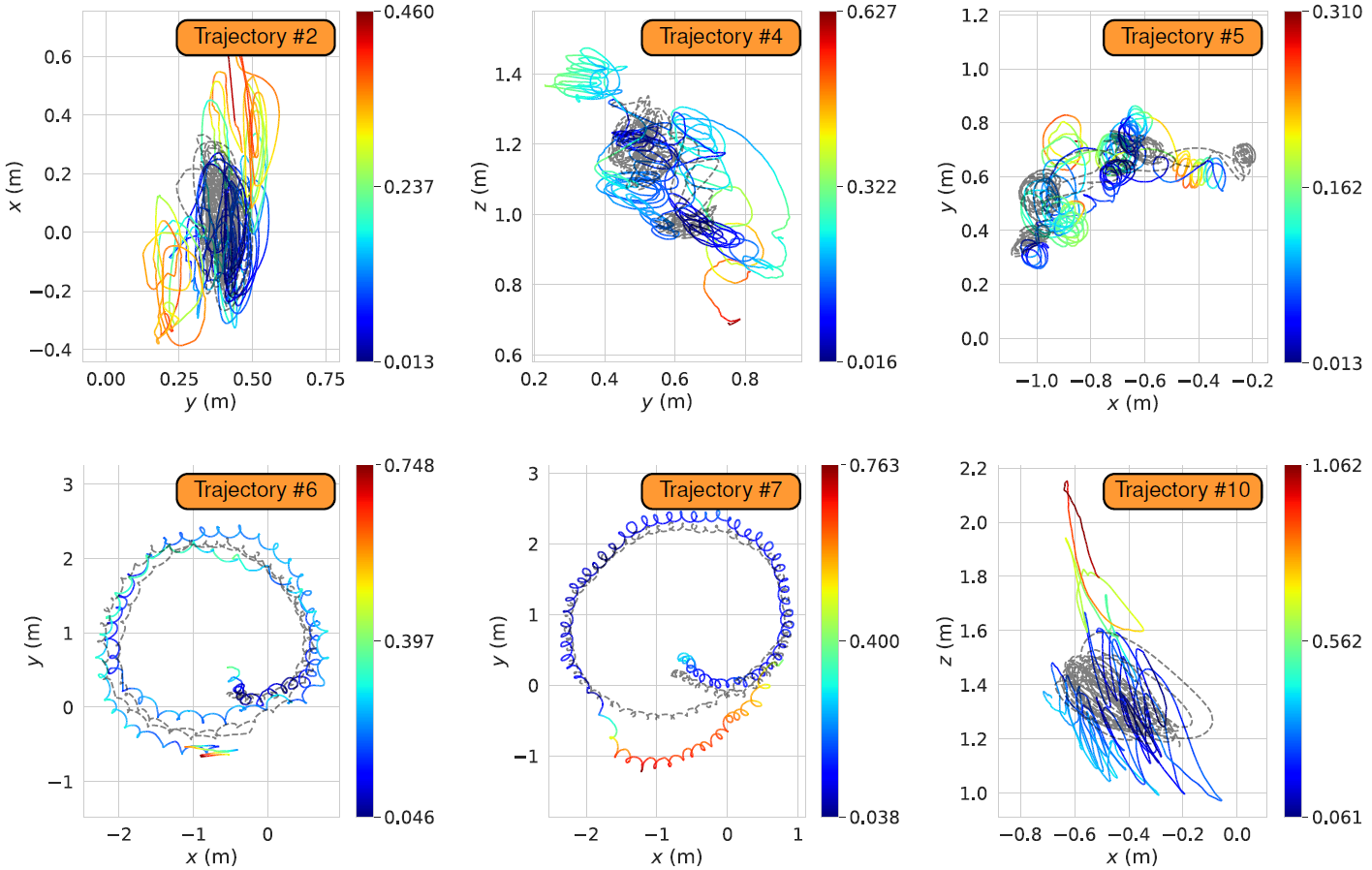}
  \vspace{-5 mm}
  \caption{Overview of representative testing trajectories used in our evaluation, aligned with the performance metrics reported in Table~\ref{tb:ate_rte}. All trajectories were executed under fast and hostile motions, as reflected by the high angular velocities (10--33 rad/s) reported in Table~\ref{tb:ate_rte}. Ground truth is shown in gray, with error mapping done for TCB-VIO.
  }
  \vspace{2 mm}
  \label{fig:traj_overview}
  \includegraphics[width=\columnwidth]{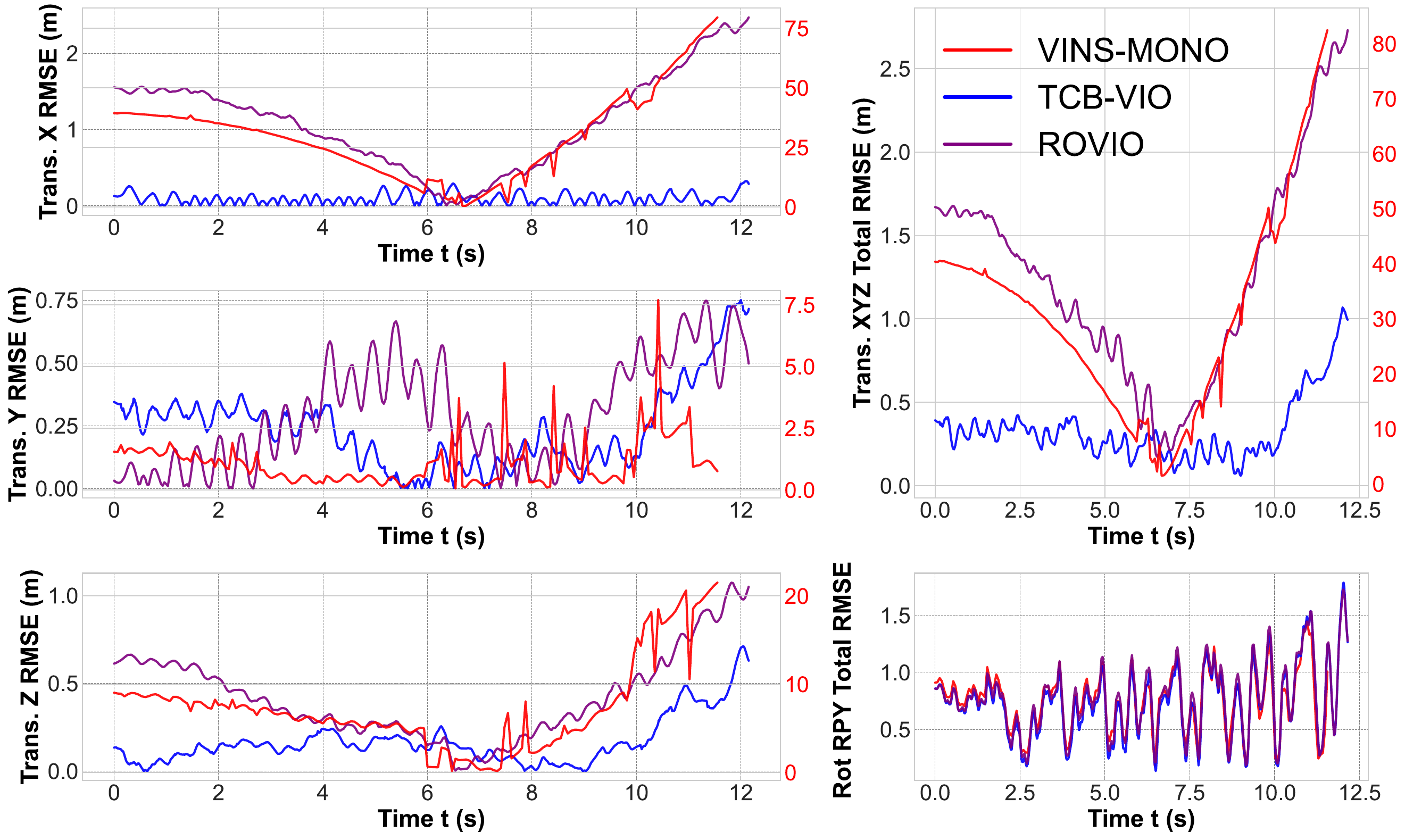}
  \vspace{-6 mm}
  \caption{Estimated RMSE for trajectory \#10: (left) translation for X, Y, and Z axes, (top right) total translation, and (bottom right) total rotation. VINS-Mono has its own axes, in red, due to large error ranges. While rotational RMSE values for all three methods are close, TCB-VIO maintains the lowest overall translational RMSE. 
  \vspace{3 mm}
}
  \label{fig:fullxyzrpy_traj2}
    \includegraphics[width=1\columnwidth]{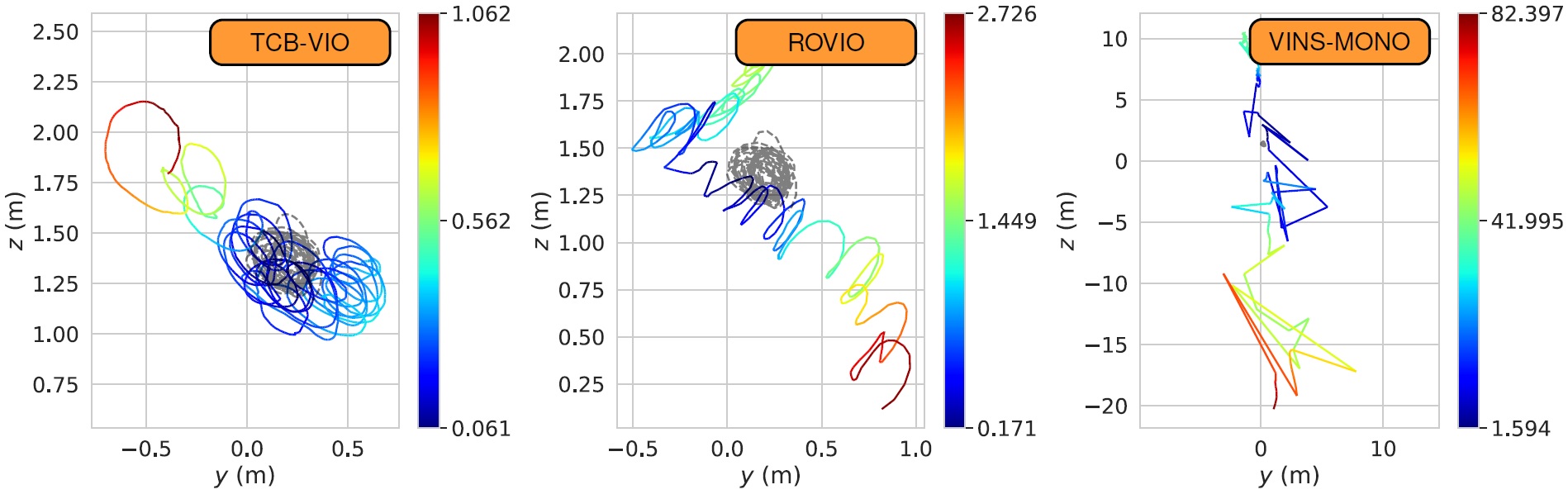}
    \vspace{-5 mm}
    \caption{Error mapping for trajectory \#10, comparing TCB-VIO (left), ROVIO (middle), and VINS-Mono (right). TCB-VIO maintains low error, \textcolor{black}{relative} to ROVIO and VINS-MONO, closely following the ground-truth path (in gray) with minimal deviations. ROVIO shows significant drift, while VINS-Mono diverges. Note that the colorbar scales differ across the graphs.}
    \label{fig:errormapping_traj2}  
\end{figure}

As an example, Fig.~\ref{fig:fullxyzrpy_traj2} shows the RMSE of trajectory \#10 along all axes. TCB-VIO achieves the lowest total translational RMSE (Fig.~\ref{fig:fullxyzrpy_traj2} top-right). 
ROVIO diverges over time, especially in the X and Z directions, with errors exceeding 2.7 meters. 
VINS-Mono has its own axes for RMSE in red on the right of subplots because of its much larger error ranges, 
particularly in the X direction, resulting in higher errors compared to TCB-VIO, with errors exceeding 30 meters in the X. These trends align with the quantitative data in Table~\ref{tb:ate_rte}, further emphasizing TCB-VIO's robustness during fast and hostile motions. 
In terms of rotational RMSE (Fig.~\ref{fig:fullxyzrpy_traj2} bottom-right), all methods perform similarly, but TCB-VIO maintains the lowest overall error. This highlights its ability to handle rapid rotational dynamics effectively, unlike ROVIO and VINS-Mono, which experience occasional spikes due to tracking loss.

Fig.~\ref{fig:errormapping_traj2} shows the tracking error, color-coded, for trajectory~\#10. \textcolor{black}{Note that the scale of the error bars is different.}  
The ground truth is shown in gray. 
TCB-VIO demonstrates 
lower error mapping 
\textcolor{black}{relative} to ROVIO and VINS-Mono algorithms.   
In contrast, both ROVIO and VINS-MONO exhibit relatively large error clusters, particularly at sharp turns and high dynamic regions, as evident from the red and orange segments on the map. Notably, VINS-MONO diverged to such an extent that it jittered and failed to maintain local consistency. 

\begin{table*}[t!]
\caption{ATE (m), first block, and RTE (m), second block, for 10 trajectories. Each cell shows RMSE/median. Lowest of all errors are in \textbf{bold}; lowest of TCB-VIO, ROVIO, VINS-Mono are \underline{underlined}. ``F'' = fail to initialize, ``PF'' = partial failure. 
Numbers after each trajectory ID indicate the rounded average angular velocity, length, and duration.
}
\centering
\scriptsize
\setlength{\tabcolsep}{3pt}
\renewcommand{\arraystretch}{1.2}
\begin{tabular*}{\textwidth}{@{\extracolsep{\fill}}|l|cccccccccc|l|}
\hline
\multirow{2}{*}{\diagbox{Alg.~\,}{Traj.}} &
\textbf{1}: {\scriptsize 11 rad/s,} & 
\textbf{2}: {\scriptsize 11 rad/s,} & 
\textbf{3}: {\scriptsize 8 rad/s} & 
\textbf{4}: {\scriptsize 10 rad/s,} & 
\textbf{5}: {\scriptsize 9 rad/s,} & 
\textbf{6}: {\scriptsize 9 rad/s,} & 
\textbf{7}: {\scriptsize 9 rad/s,} &
\textbf{8}: {\scriptsize 13 rad/s,} & 
\textbf{9}: {\scriptsize 14 rad/s,} & 
\textbf{10}: {\scriptsize 15 rad/s,} & \\

& 
{\scriptsize {25 m, 19 s}} & 
{\scriptsize {33 m, 26 s}} & 
{\scriptsize {24 m, 29 s}} &  
{\scriptsize {20 m, 24 s}} &  
{\scriptsize {21 m, 26 s}} & 
{\scriptsize {29 m, 24 s}} & 
{\scriptsize {31 m, 31 s}} & 
{\scriptsize {26 m, 16 s}} & 
{\scriptsize {28 m, 13 s}} & 
{\scriptsize {25 m, 12 s}} & 
\multirow{2}{*}{} \\ \hline

TCB-VIO &
\underline{0.199}/\underline{0.202} &
\underline{0.223}/\underline{0.156} &
0.158/0.113 &
\textbf{\underline{0.218}}/\textbf{\underline{0.192}} &
\textbf{\underline{0.140}}/\textbf{\underline{0.121}} &
\textbf{\underline{0.255}}/\textbf{\underline{0.234}} &
\textbf{\underline{0.315}}/\textbf{\underline{0.168}} &
\textbf{\underline{0.210}}/\textbf{\underline{0.136}} &
\textbf{\underline{0.601}}/\textbf{\underline{0.469}} &
\textbf{\underline{0.379}}/\textbf{\underline{0.013}} & A \\

Ablation 1 &
0.354/0.207 &
10.056/5.666 &
\textbf{\underline{0.107}}/\textbf{\underline{0.073}} &
F/F &
PF/PF &
F/F &
PF/PF &
F/F &
PF/PF &
F/F & T \\

Ablation 2 &
PF/PF & PF/PF & F/F & PF/PF & 5.778/3.386 & PF/PF & 1.375/1.152 & PF/PF & F/F & F/F & E \\

ROVIO &
0.342/0.220 &
0.693/0.433 &
0.309/0.167 &
2.548/1.854 &
9.241/6.126 &
3.832/3.378 &
PF/PF &
0.648/0.629 &
6.501/5.415 &
1.405/1.245 & \\

VINS-Mono &
0.113/0.083 &
0.563/0.416 &
2.798/1.203 &
PF/PF &
PF/PF &
PF/PF &
PF/PF &
PF/PF &
PF/PF &
PF/PF  &\\

ORB-SLAM3 &
\textbf{0.091}/\textbf{0.053} &
\textbf{0.151}/\textbf{0.096} &
PF/PF &
PF/PF &
PF/PF &
F/F &
F/F &
F/F &
F/F &
F/F & \\
\hline

TCB-VIO &
\textbf{\underline{0.021}}/\textbf{\underline{0.008}} &
\textbf{\underline{0.019}}/\textbf{\underline{0.010}} &
\textbf{\underline{0.010}}/\textbf{\underline{0.003}} &
\textbf{\underline{0.009}}/\textbf{\underline{0.004}} &
\textbf{\underline{0.009}}/\textbf{\underline{0.005}} &
\textbf{\underline{0.016}}/\textbf{\underline{0.009}} &
\textbf{\underline{0.011}}/\textbf{\underline{0.008}} &
\textbf{\underline{0.024}}/\textbf{\underline{0.013}} &
\textbf{\underline{0.015}}/\textbf{\underline{0.011}} &
\textbf{\underline{0.028}}/\textbf{\underline{0.009}} &R \\

Ablation 1 &
0.052/0.036 &
0.085/0.051 &
0.035/0.024 &
F/F &
PF/PF &
F/F &
PF/PF &
F/F &
PF/PF &
F/F &  T\\

Ablation 2 &
PF/PF & PF/PF & F/F & PF/PF & 0.013/0.007 & PF/PF & 0.013/0.010 & PF/PF & F/F & F/F &  E\\

ROVIO &
0.085/0.062 &
0.078/0.067 &
0.054/0.040 &
0.051/0.042 &
0.079/0.056 &
0.076/0.058 &
PF/PF &
0.089/0.083 &
0.149/0.118 &
0.112/0.105 & \\

VINS-Mono &
0.171/0.149 &
0.178/0.140 &
0.202/0.093 &
PF/PF &
PF/PF &
PF/PF &
PF/PF &
PF/PF &
PF/PF &
PF/PF & \\

ORB-SLAM3 &
0.091/0.072 &
0.098/0.071 &
PF/PF &
F/F &
PF/PF &
F/F &
F/F &
F/F &
F/F &
F/F & \\
\hline
\end{tabular*}
\label{tb:ate_rte}
\end{table*}
\begin{figure*}[htbp!]
\centerline{
\includegraphics[width =\textwidth]{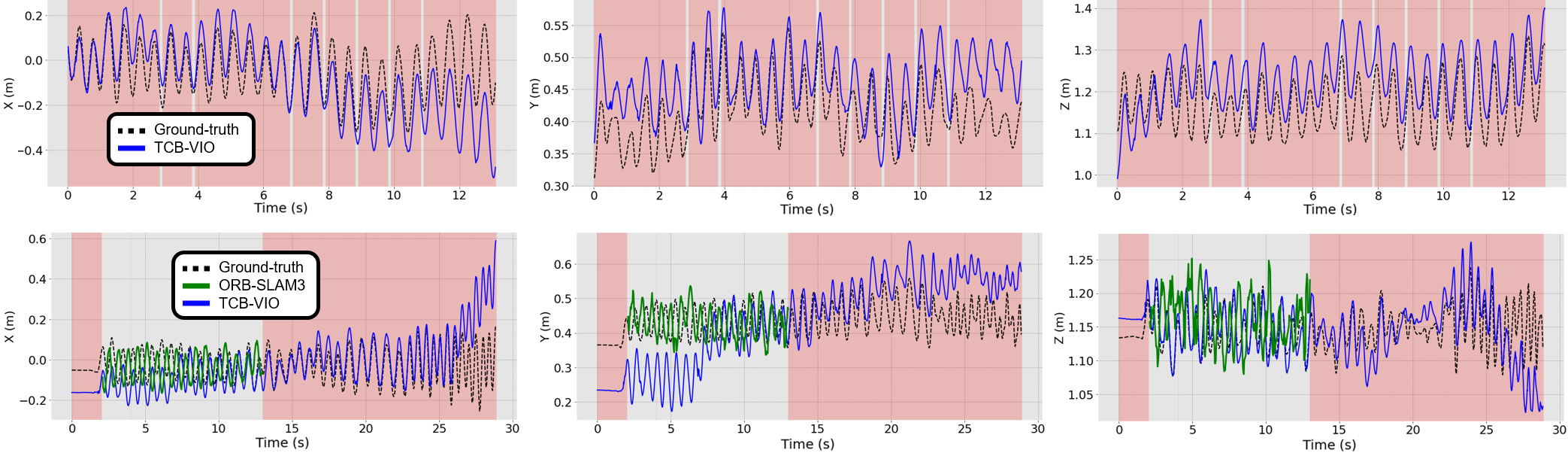}
}
\caption{
(Top row) Trajectory \#8 (F-type failure) showing failure during pink regions. TCB-VIO tracks consistently despite occasional deviations from ground-truth. 
%
(Bottom row) Trajectory \#3 (PF-type failure) illustrating ORB-SLAM3 reinitializing but quickly diverging, highlighted in pink, leading to ATE and RTE errors.
}
\label{fig:orbslam3comparison2x3}
\end{figure*}

\subsection{Trajectory Failures Present in ORB-SLAM3}
As shown in Table~\ref{tb:ate_rte}, while the metrics of TCB-VIO and ORB-SLAM3 for trajectories \#1 and \#2 are close, ORB-SLAM3 fails in the rest of the trajectories. In the table, an ``F'' indicates a failure to initialize, and ``PF'' is a partial failure.
Fig.~\ref{fig:orbslam3comparison2x3}-(top row) corresponds to trajectory \#8, where ORB-SLAM3 fails. This type of failure is characterized by ORB-SLAM3's inability to initialize in certain regions, specifically those marked in pink. In the short non-pink regions, ORB-SLAM3 performs well, demonstrating its potential under favorable conditions. However, during pink intervals, ORB-SLAM3 crashes, losing tracking completely. In contrast, TCB-VIO, while deviating slightly from the ground truth in some areas, maintains reasonable tracking accuracy and does not suffer from the catastrophic failures observed in ORB-SLAM3.

Fig.~\ref{fig:orbslam3comparison2x3}-(bottom row) illustrates a PF-type failure, as seen in trajectory \#3 from Table~\ref{tb:ate_rte}. In this particular trajectory, compared to trajectories \#1-2, the scene was designed to contain fewer trackable features, making successful tracking possible primarily in high-exposure conditions where the binarized image remained stable.
Under these constraints, ORB-SLAM3 repeatedly attempts to reinitialize within the pink regions but diverges rapidly, leading to large ATE and RTE values. Although the pink regions are less prominent compared to F-type failures, their impact is still significant due to the rapid loss of tracking. 
The limited feature availability, coupled with faster motions, caused ORB-SLAM3’s failure to recover. In contrast, 
TCB-VIO demonstrates its robustness by maintaining a consistent trajectory, mitigating errors, and maintaining reliable performance throughout the trajectory, despite the challenging conditions.

\subsection{Discussions}
This section discusses five aspects of TCB-VIO's performance and design, including its 1) time complexity, 2) computational load, 3) the impact of in-state (SLAM) features, 4) outdoor performance, and 5) ablation studies.
\subsubsection{Time Complexity Analysis}\label{sec:time_complexity}
To quantify the computational benefits of running TCB-VIO on the FPSP, the execution times of edge and corner detection on the FPSP were compared against CPU and GPU. 
For the CPU/GPU measurements, 1000 images were captured by the SCAMP-5 FPSP and processed to perform edge and corner detection. Average per-frame results were reported using system utility tools. For FPSP, execution times were calculated from the number of instructions in the code and the execution time per instruction.

As shown in Table~\ref{tab:sobel_runtime}, for edge detection, the SCAMP-5 FPSP achieves $16.4\times$ speedup over the CPU and $17.0\times$ over the GPU. For feature extraction (1000 keypoints), SCAMP-5 FPSP requires 292 $\mu s$, while CPU and GPU require 76.67 $\mu s$ and 20.23 $\mu s$. On SCAMP, the corner detection runtime is constant regardless of keypoint count since all $256\times256$ processing elements execute neighbour comparisons in parallel. Thus, the cost per frame is fixed. In contrast, CPU runtimes scale with keypoints. SCAMP-5 may appear slower in Table~\ref{tab:sobel_runtime}, but this reflects its fixed cost model rather than inefficiency. SCAMP-5 executes operations at 5-10 MHz, far below conventional processors (CPU $\approx2.7$ GHz, GPU $\approx400$ MHz). This frequency gap explains SCAMP's longer absolute runtimes despite its parallel architecture. Future FPSP designs with higher clock rates would narrow this gap while preserving parallelism. Note that the intensity image transfer time, significant for CPU/GPU, was not included. By avoiding intensity image transfer, host processor saturation is prevented.
\begin{table}[h!]
\scriptsize
\setlength{\tabcolsep}{3pt}
\renewcommand{\arraystretch}{1.2}
\centering
\vspace{-5 mm}
\caption{Runtime comparison on FPSP/CPU/GPU. }
\begin{tabular}{c|c|c|c}
\hline
Process & FPSP (SCAMP-5)  & CPU (i7- 12650H) & GPU (RTX 3070)\\
\hline
Edge Detection  & {9.0} $\mu s$ & {147.818} $\mu s$ & {152.63} $\mu s$\\
\hline
Corner Detection  & {292.0} $\mu s$ & {76.671} $\mu s$ & {20.23} $\mu s$\\
\hline
\end{tabular}
\label{tab:sobel_runtime}
\end{table}
\subsubsection{Computational Load Analysis}\label{sec:comp_load}
The efficiency of TCB-VIO was evaluated in terms of energy consumption. Edge and corner detection were executed directly on-sensor by the SCAMP-5 FPSP using its dedicated processing elements, resulting in per-frame energy consumption of only tens of millijoules (mJ). In contrast, substantially higher CPU/GPU utilization was required when the same operations were offloaded to conventional processors. 
Per-frame energy consumption was measured for edge and corner detection on the FPSP, CPU, and GPU. For CPU/GPU evaluation, 1000 images were captured by the SCAMP-5 FPSP while the sensor was moved along a known trajectory. For the FPSP evaluation, the same trajectory was repeated twice, once with edge detection and once with corner detection enabled. 
Energy measurements were obtained using a USB power meter for the SCAMP-5 FPSP, while Intel RAPL and Nvidia SMI were employed to monitor CPU and GPU energy consumption, respectively.

Table~\ref{tab:sobel_load} reports the per-frame energy consumption for the Sobel edge detection (kernel size  $k=3$) and BIT corner detection, averaged on 1000 $256 \times 256$ images. For edge detection, SCAMP-5 consumes only $\sim${0.019 mJ} per frame, whereas the CPU/GPU requires {994.23 mJ/6.46 mJ} per frame. 
For corner detection, the FPSP consumes {1.15 mJ} per frame, compared to {987.22 mJ/1.65 mJ} on the CPU/GPU. This computational load analysis confirms that SCAMP-5 reduces energy consumption relative to a CPU/GPU-based pipeline, underscoring the lightweight and embedded nature of on-sensor SIMD computation.
\begin{table}[h!]
\scriptsize
\setlength{\tabcolsep}{3pt}
\renewcommand{\arraystretch}{1.2}
\centering
\caption{Average per-frame energy consumption on FPSP/CPU/GPU. }
\begin{tabular}{c|c|c|c}
\hline
Process & FPSP (SCAMP-5)  & CPU (i7- 12650H) & GPU (RTX 3070)\\
\hline
Edge Detection  & {0.019} mJ & 994.23 mJ & {6.46 mJ}\\
\hline
Corner Detection  & {1.15} mJ & 987.22 mJ & {1.65 mJ}\\
\hline
\end{tabular}
\label{tab:sobel_load}
\end{table}

\subsubsection{Impact of in-state Features}\label{sec:deviation}
\begin{figure}[h!]
    \centering
    \includegraphics[width=\linewidth]{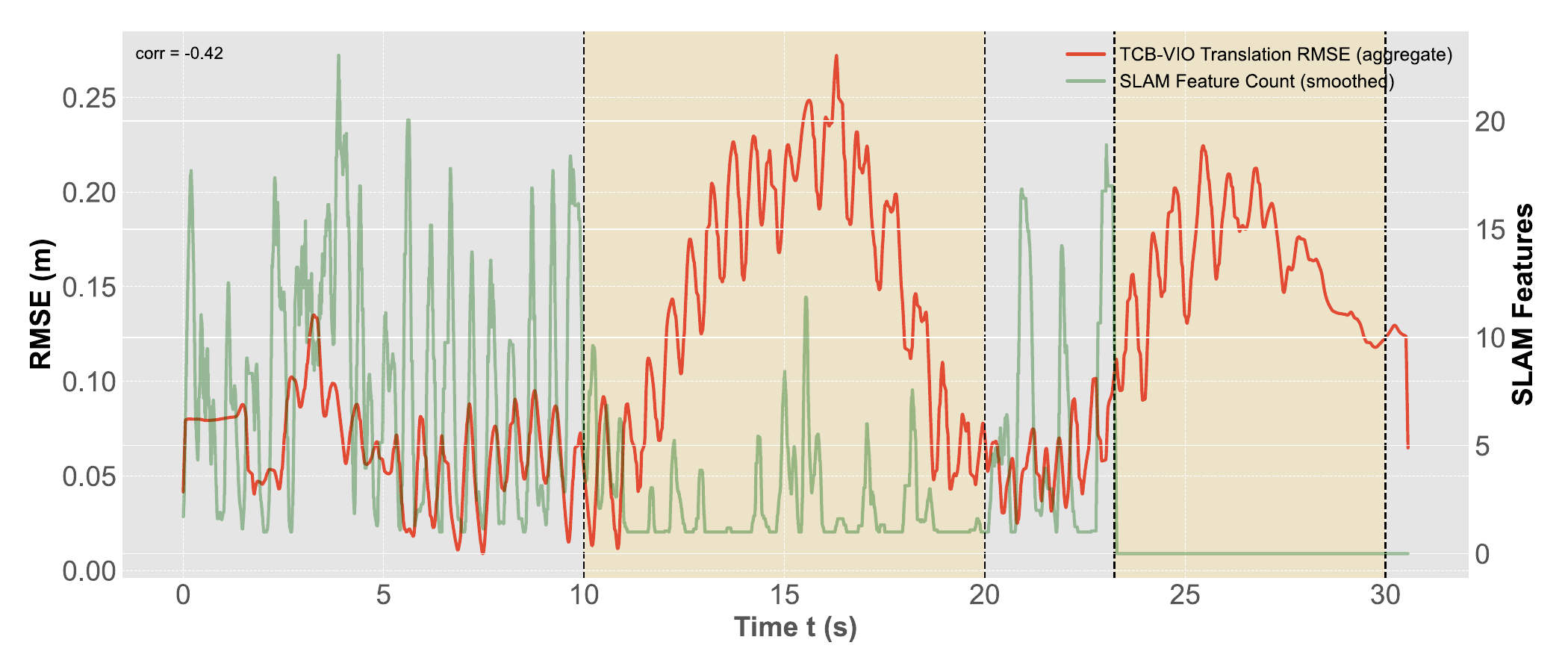}
    \vspace{-8 mm}
    \caption{Correlation between in-state feature and translational RMSE for trajectory~\#2. When feature count is <10 (yellow regions), TCB-VIO has higher translational RMSE.}
    \label{fig:error-cause1}
\end{figure}

TCB-VIO consistently achieves low ATE and RTE, but deviations from ground truth occur under specific conditions. Fig.~\ref{fig:error-cause1} shows RMSE translation error (deviation) and in-state feature count, for trajectory~\#2, overlaid. 
In yellow regions, due to poor texture, feature count is low and RMSE is high, the converse is true in green regions. 
The negative correlation ($\text{corr}=-0.42$) shows that having persistent features helps constrain long-term drift. This trend was seen in other trajectories as well.
\subsubsection{Outdoor Trajectories}\label{sec:outdoor}
In addition to the 10 indoor experiments, three additional experiments were conducted outdoors. Lacking ground truth in the outdoor environment, the performance was evaluated by starting and ending the trajectories from a single fixture, to ensure the start and end poses were identical. The VIO performance was then measured by the endpoint drift (distance) between the estimated start and end poses. As indicated in Table~\ref{tb:start_to_end}, TCB-VIO exhibits the lowest error compared to the other algorithms. Videos of these experiments are on the project website~\cite{web:tcbvio}.
\begin{table}[h!]
\caption{Start-to-end error (m) for outdoor experiments, performed in sun/shade contrast. Lowest error is in \textbf{bold}. ``F'' and ``PF'' are for `fail to initialize' and `partial failure'. The last column shows the average angular velocity and trajectory length/duration. The velocity and length have been computed using the TCB-VIO pose estimates.}
\centering
\scriptsize
\setlength{\tabcolsep}{2pt} 
\renewcommand{\arraystretch}{1.2}
\begin{tabular}{c|cccc|c} 
\hline
Traj.&
TCB-VIO & 
ROVIO & 
VINS-Mono & 
ORB-SLAM3  &
Description of Trajectory \\ \hline

11  & 
\textbf{\underline{1.713}} & 
21.6 (PF) & 
F & 
F & 
7 rad/s,  21 m, 20.3 s \\

12  & 
\textbf{\underline{0.130}} & 
11.8 (PF) & 
695.5 (PF) & 
F & 
9 rad/s,  47 m, 49.9 s \\

13  & 
\textbf{\underline{2.736}} & 
5.1 & 
F & 
F & 
8 rad/s,  31 m, 15.9 s \\ \hline
\end{tabular}
\label{tb:start_to_end}
\end{table}
\subsubsection{Ablation Studies}\label{sec:abalation}
To evaluate the design of the algorithm, two ablation studies were conducted, as summarized in Table~\ref{tb:ate_rte}. The corresponding ATE and RTE results are reported in the rows labelled Ablation 1 and Ablation 2. 
In Ablation 1, edge images were transmitted to the host, and Shi–Tomasi features (extracted from the edges) were used for KLT tracking instead of BIT features. The only difference from the TCB-VIO pipeline is the type of features. 
In Ablation 2, the feathering block (Sec.~\ref{sec:feather}) was removed from the TCB-VIO pipeline.

Table~\ref{tb:ate_rte} shows that across all trajectories, TCB-VIO achieves lower ATE and RTE than the ablation studies, with the exception of trajectory~\#3, where its performance is nearly identical to Ablation~1. For instance, in trajectory~\#1, the median ATE decreases from 0.207m (Ablation~1) to 0.202m (TCB-VIO). More importantly, in challenging sequences such as trajectory~\#9, Ablation~1 fails to track, while Ablation 2 fails to initialize. These results highlight that both the choice of features and the inclusion of the feathering process are essential for the effectiveness of the TCB-VIO pipeline.

\textcolor{black}{The established benchmark algorithms are known to perform well under standard and low-speed conditions. Our evaluation is specifically designed to highlight performance gains beyond the operational limits of conventional methods, which is the main contribution of TCB-VIO.}

\section{Conclusion}\label{sec:conc}
In this work, we presented TCB-VIO, a tightly-coupled 6-DoF VIO framework using the on-sensor compute capabilities of SCAMP5 FPSP. By integrating visual and inertial measurements at 250 FPS and 400 Hz, respectively, and processing binary edge images and feature maps with a novel binary-enhanced Kanade-Lucas-Tomasi (KLT) tracker, TCB-VIO achieves robust and accurate tracking under fast motions. 
The evaluations demonstrated that TCB-VIO outperforms state-of-the-art methods, including ROVIO, VINS-Mono, and ORB-SLAM3 in terms of both translational and rotational accuracy. While ROVIO exhibits drift over time, and VINS-Mono fails to maintain tracking under extreme conditions, TCB-VIO maintains accurate trajectory estimation. This highlights the importance of high frame-rate synchronization between visual and inertial data, as well as the computational advantages provided by FPSPs. 
The proposed framework addresses the trade-off between tracking performance and latency by combining high-frequency visual-inertial data processing with efficient on-sensor computation. This work demonstrates the potential of FPSPs in advancing the capabilities of mobile robotic systems, paving the way for more reliable visual-inertial odometry. 

The proposed method has partial on-sensor computation due to hardware limitations. 
Future hardware with more flexible compute units could enable all computations to be migrated entirely to the FPSP. In particular, while the feathering process is highly parallelizable, it demands significant bandwidth to transfer the results to the host. An optimal solution would shift this computational load from the host to the FPSP. Another approach is to utilize lower-dimensional and more computationally efficient features for tracking; the corner and edge features in the binary-enhanced KLT algorithm in this work show promise for further investigation.

{\small
\bibliographystyle{ieeetran}
\bibliography{sample}
}

\end{document}